\documentclass{article}

\usepackage{arxiv}

\usepackage[utf8]{inputenc} 
\usepackage[T1]{fontenc}    
\usepackage{hyperref}       
\usepackage{url}            
\usepackage{booktabs}       
\usepackage{amsfonts}       
\usepackage{nicefrac}       
\usepackage{microtype}      
\usepackage{lipsum}
\usepackage{graphicx}
\usepackage{amsmath}
\usepackage{multicol}
\usepackage{multirow}
\usepackage{slashbox}

\title{2D Image Features Detector And Descriptor Selection Expert System}

\author{
 Ibon Merino \\
  Industry and Transport\\ 
  Tecnalia Research and Innovation\\ 
  Donostia-San Sebastian\\
  Spain\\
  \texttt{ibon.merino@tecnalia.com} \\
   \And
 Jon Azpiazu \\
  Industry and Transport\\ 
  Tecnalia Research and Innovation\\ 
  Donostia-San Sebastian\\
  Spain\\
  \texttt{jon.azpiazu@tecnalia.com} \\
  \And
 Anthony Remazeilles \\
  Industry and Transport\\ 
  Tecnalia Research and Innovation\\ 
  Donostia-San Sebastian\\
  Spain\\
  \texttt{anthony.remazeilles@tecnalia.com} \\
  \And
 Basilio Sierra \\
  Computer Science and Artificial Intelligence\\ 
  University of the Basque Country UPV/EHU\\ 
  Donostia-San Sebastian\\
  Spain
  \texttt{b.sierra@ehu.eus} \\
}

\begin{document}
\maketitle
\begin{abstract}
Detection and description of keypoints from an image is a well-studied problem in Computer Vision. Some methods like SIFT, SURF or ORB are computationally really efficient. This paper proposes a solution for a particular case study on object recognition of industrial parts based on hierarchical classification. Reducing the number of instances leads to better performance, indeed, that is what the use of the hierarchical classification is looking for. We demonstrate that this method performs better than using just one method like ORB, SIFT or FREAK, despite being fairly slower. 
\end{abstract}


\section{Introduction \label{sec:introduction}}
Object recognition is an important branch of computer vision. Its main idea is to extract important data or features from images in order to recognize which object is present on it. Many different techniques are used in order to achieve this. In recent computer vision literature, it has been a widely spread tendency to use deep learning due to their benefits throwing out many techniques of previous literature that, actually, have a good performance in many cases. Our aim is to recover those techniques in order to boost them and increase their performance or use their benefits that neural networks may not have.

The classical methods in computer vision are based in pure mathematical operations were images are used as matrices. These methods look for gradient changes, patterns... and try to find similarities in different images or build a machine learning model to try to predict the objects that are present in the image. 

Our use case is the industrial area were many similar parts are to be recognized. Those parts vary a lot from one to another (textures, size, color, reflections,...) so an expert is needed for choosing which method is better for recognizing the objects. We propose a method that simulates the expert role. This is achieved learning a model that classifies the objects in groups that behave similarly to different recognition methods. This leads to a hierarchical classification that first classifies the object to be recognized in one of the previously obtained groups and inside the group the method that works better in that group is used to recognize the object. 

The paper is organized as follows. In Section~\ref{sec:background} we present a state of art of the most used 2D feature-based methods, including detectors, descriptors and matchers. The purpose of Section \ref{sec:proposed-approach} is to present the method that we propose and how we evaluate it. The experiments done and their results are shown in section \ref{sec:experiment}. Section \ref{sec:conclusion} summarizes the conclusions that can be drawn from our work.
 
\section{Background \label{sec:background}}
There are several methods for object recognition. In our case, we have focused on feature-based methods. These methods look for points of interest of the images (detectors), try to describe them (descriptors) and match them (matchers). The combination of different detectors, descriptors and matchers vary the perfomance of the whole system. This is a fast growing area in image processing field. The following short and chronologically ordered review presents the gradual improvements in feature detection (Subsection \ref{subsec:detectors}), description (Subsection \ref{subsec:descriptors}) and matching (Subsection \ref{subsec:matchers}). 

\subsection{2D features detectors \label{subsec:detectors}}
One of the most used methods was proposed in 1999 by \cite{lowe_object_1999}. This method is called SIFT, which stands for Scale Invariant Feature Transform. The main idea is to use the Difference-of-Gaussian function (a close approximation to the Laplacian-of-Gaussian proposed by Lowe) to search for extrema in the scale space. Even if SIFT was relatively fast, a new method, SURF (Speeded Up Robust Features) \cite{leonardis_surf:_2006}, outperforms it in terms of repeatability, distinctiveness and robustness, although it can be computed and compared much faster.

In addition, FAST (Features from Accelerated Segment Test) proposed by \cite{rosten_fusing_2005} introduce a fast detector. FAST outperforms previous algorithms (like SURF and SIFT) in both computational performance and repeatability. AGAST \cite{mair2010adaptive} is based on the FAST, but it is more efficient as well as generic. BRISK \cite{leutenegger_brisk:_2011} is a novel method for keypoint detection, description and matching which has a low computational cost (as stated in the corresponding article, an order of magnitude faster than SURF in some cases). Following the same line of FAST based mehods, we find ORB \cite{rublee_orb:_2011}, an efficient alternative to SIFT or SURF. This method's detector is based on FAST but it adds orientation in order to obtain better results. In fact, this method performs at two orders of magnitude faster than SIFT, in many situations. 

\subsection{2D features descriptors \label{subsec:descriptors}}
\cite{lowe_object_1999} also proposed a descriptor called SIFT. As mentioned above, is one of the most popular feature detector and descriptor. The descriptor is a position-dependent histogram of local image gradient directions around the interest point and is also scale invariant. It has numerous extensions such as PCA-SIFT \cite{yan_ke_pca-sift:_2004}, that mixes PCA with SIFT; CSIFT \cite{abdel2006csift}, Color invariant SIFT; GLOH \cite{mikolajczyk_performance_2005}; DAISY \cite{tola_daisy:_2010}, a dense descriptor inspired in SIFT and GLOH; and so on. SURF descriptor \cite{leonardis_surf:_2006} relies on integral images for image convolutions in order to obtain its speed.

BRIEF \cite{calonder2010brief} is a highly discriminative feature descriptor that is fast both to build and to match. BRISK \cite{leutenegger_brisk:_2011} descriptor is composed as a binary string by concatenating the results of simple brightness comparison tests. ORB descriptor is BRIEF-based and adds rotation invariance and resistance to noise.

LBP (Local Binary Patterns) \cite{ojala_comparative_1996} is a two-level version of the texture spectrum method \cite{wang1990texture}. This methods has been really popular and many derivatives has been proposed. Based on this, the CS-LBP (Center-Symmetric Local Binary Pattern) \cite{heikkila_description_2009} combines the strengths of SIFT and LBP. Later in 2010, the LTP (Local Ternary Pattern) \cite{liao_region_2010} appeared, a generalization of the LBP that is more discriminant and less sensitive to noise in uniform regions. Same year, ELTP (Extended local ternary pattern) \cite{nanni2010local} improved this by attempting to strike a balance by using a clustering method to group the patterns in a meaningful way. In 2012, LTrP (Local Tetra Patterns)  \cite{murala_local_2012} encoded the relationship between the referenced pixel and its neighbors, based on the directions that are calculated using the first-order derivatives in vertical and horizontal directions. In \cite{pietikainen_local_2011} there are gathered other methods that are based on the LBP.

Other descriptor called FREAK \cite{alahi_freak:_2012} is a keypoint descriptor inspired by the human visual system and more precisely the retina. It is faster, usess less memory and more robust than SIFT, SURF and BRISK. They are thus competitive alternatives to existing descriptors in particular for embedded applications.

\subsection{Matchers \label{subsec:matchers}}
The most widely used method for matching is Nearest Neighbor (NN). Many algorithms follow this method. One of the most used is the kd-tree \cite{Robinson:1981:KSS:582318.582321} which works well with low dimensionality. For dealing with higher dimensionalities many researchers have proposed diverse methods such as the Approximate Nearest Neighbor (ANN) by \cite{Indyk:1998:ANN:276698.276876} or the Fast Approximate Nearest Neighbors of \cite{muja2009fast} which is implemented in the well known open source library FLANN (Fast Library for Approximate Nearest Neighbors).

\section{Proposed Approach \label{sec:proposed-approach}}
As we have stated before, the issue we are dealing with is the recognition of industrial parts for pick-and-placing. The main problem is that the accurate recognition of some kind of parts are highly dependant on the recognition pipeline used. This is because parts' characteristics like texture (presence or absence), forms, colors, brightness; make some detectors or descriptors work differently. We are thus proposing a systematic approach for selecting the best recognition pipeline for a given object (Subsection \ref{subsec:RecognitionEvaluation}). We also propose in Subsection \ref{subsec:ExpertSystem} an expert system that identifies groups of parts that are recognized similarly to improve the overall accuracy. The recognition pipeline is explained in Subsection \ref{subsec:RecognitionPineline}.\\

We start defining some notations. An industrial part, or object, is named \textbf{instance}. The images captured of each part are named \textbf{views}. Given the set of views $\boldsymbol{X}$, the set of instance labels $\boldsymbol{Y}$ and the set of recognition pipelines $\boldsymbol{\Psi}$, the function $\omega_{X, Y}^{\Psi}(y)$ returns for each $y\in\boldsymbol{Y}$ the best pipeline $\psi * \in \boldsymbol{\Psi}$ according to a metric $F_1$ that is later discussed. We call $\psi^{**}$ to the pipeline that on average performs better according to the evaluation metric, this is, that maximizes the average of the scores per instance (\ref{eq:psi*}).

\begin{equation}
\label{eq:F1_intra}
\omega_{X, Y}^{\Psi}(y)= \displaystyle \underset{\psi \in \Psi}{\mathrm{argmax}}\,{F_1}_{y}^{\psi}(X,Y)=\psi^*
\end{equation}

\begin{equation}
\label{eq:psi*}
\psi^{**}= \underset{\psi \in \Psi}{\mathrm{argmax}}\,\frac{\displaystyle \sum_{y\in Y} {F_1}_y^{\psi}(X,Y)}{|Y|}
\end{equation}

\subsection{Recognition Pipeline \label{subsec:RecognitionPineline}}

\begin{figure*}
    \centering
    \includegraphics[width=\textwidth]{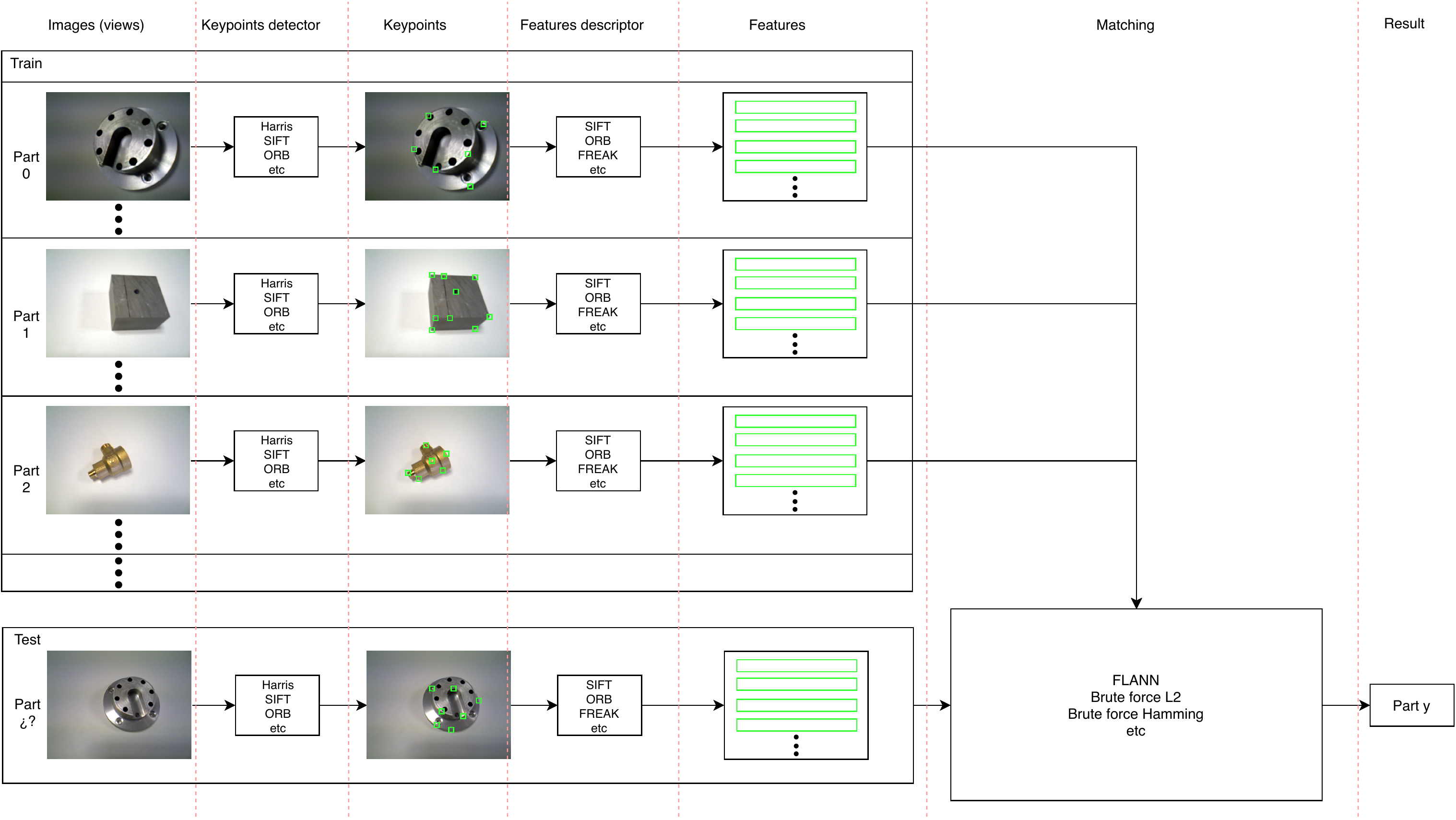}
    \caption{Recognition pipeline}
    \label{fig:recognition_pipeline}
\end{figure*}

A recognition pipeline $\Psi$ is composed of 3 steps: detection, description and matching. Detectors, $\boldsymbol{\Gamma}$, localize interesting keypoints in the view (gradient changes, changes in illumination,...). Descriptors, $\boldsymbol{\Phi}$, are used to represent those keypoints in order to locate them in other views. Matchers, $\boldsymbol{\Omega}$, find the closest features between views. So, a pipeline $\psi$ is composed by a keypoint detector $\gamma$, a feature descriptor $\phi$ and a matcher $\omega$. Figure \ref{fig:recognition_pipeline} shows the structure of the recognition pipeline.

The keypoints detection and description are described previously in the background section. In the matching, are two groups of features: the ones that form the model (train) and the ones that need to be recognized (test). Different kind of methods could be used to match features, but, mainly, distance based techniques are used. This techniques make use of different distances (L2, hamming,...) to find the closest feature to the one that needs to be labeled. Those two features (the test feature and the closest to this one) are considered a match. 
In order to discard ambiguous features, we use the Lowe's ratio test \cite{lowe_distinctive_2004} to define whether two features are a "good match". Assuming $f_t$ is the feature to be recognized, and  ${f_l}_1$ and ${f_l}_2$ its two closest features from the model, then ($f_t$, ${f_l}_1$) is a good match if:
\begin{equation}
\label{eq:lowe}
    \frac{d(f_t,f_{l_1})}{d(f_t,f_{l_2})}<r
\end{equation}

where $d(f_A, f_B)$ is the distance (L2, Hamming,...) between features A and B, and $r$ is a threshold that is used to validate if two features are similarly close to the test feature and discard it. This threshold is set at $0.8$. Now a simple voting system is used for labeling the view. For each view from the model (train) the number of good matches are counted. The good matches of each instances are summed and the test view is labeled as the instance with more good matches. 

\subsection{Recognition Evaluation \label{subsec:RecognitionEvaluation}}

As we have said, we have the input views $X$, the instance labels $Y$ and the pipelines $\Psi$. To evaluate the pipelines we have to separate the views in train and test. The evaluation method used for it is Leave-One-Out Cross-Validation (LOOCV) \cite{kohavi1995study}. It consists of $|X|$ iterations, that for each iteration $i$, the train dataset is $(X-x_i)$ and the test sample is $x_i$. With this separation train-test we can generate the confusion matrix. 
Table \ref{table:confusion_matrix} is an example of a confusion matrix for 3 instances.

\begin{table}[]
\centering
\caption{Example of a confusion matrix for 3 instances.}
\label{table:confusion_matrix}
\begin{tabular}{cc|c|c|c|c}
\cline{3-5}
\multicolumn{2}{c|}{\multirow{2}{*}{}}                                                                          & \multicolumn{3}{c|}{Actual instance} &                          \\ \cline{3-5}
\multicolumn{2}{c|}{}                                                                                           & object 1   & object 2   & object 3   &                          \\ \hline
\multicolumn{1}{|c|}{\multirow{3}{*}{\begin{tabular}[c]{@{}c@{}}Predicted\\ instance\end{tabular}}} & object 1 & 40         & 10         & 0          & \multicolumn{1}{c|}{50}  \\ \cline{2-6} 
\multicolumn{1}{|c|}{}                                                                              & object 2 & 0          & 30         & 25         & \multicolumn{1}{c|}{55}  \\ \cline{2-6} 
\multicolumn{1}{|c|}{}                                                                              & object 3 & 10         & 10         & 25         & \multicolumn{1}{c|}{45}  \\ \hline
                                                                                                    &          & 50         & 50         & 50         & \multicolumn{1}{c|}{150} \\ \cline{3-6} 
\end{tabular}

\end{table}

As mentioned in the introduction of Section \ref{sec:proposed-approach}, we use the metric $F_1$ value \cite{goutte2005probabilistic} for scoring the performance of the system. The score is calculated for the tests views from the LOOCV. $F_1$ score, or value, is calculated per each instance (\ref{eq:F1_per_instance}). This metric is an harmonic mean between the precision and the recall. The mean of all the $F_1$'s,\, $\bar{F_1}$ (\ref{eq:F1_mean}) is used for calculating $\psi^{**}$.

\begin{equation}
\label{eq:F1_per_instance}
F_1(y) = 2\cdot\frac{precision_{y} * recall_{y}}{precision_{y} + recall_{y}}
\end{equation}

\begin{equation}
\label{eq:F1_mean}
\bar{F_{1}} = \frac{\sum_{y \in Y} F_1(y)}{|Y|}
\end{equation}

The precision (Equation \ref{eq:precision}) is the ratio between the correctly predicted views with label $y$ ($tp_{y}$) and all predicted views for that given instance ($|\psi(X)=y|$). The recall (Equation \ref{eq:recall}), instead, is the relation between correctly predicted views with label $y$ ($tp_{y}$) and all views that should have that label ($|label(X)=y|$).

\begin{equation}
\label{eq:precision}
precision_{y} = \frac{tp_{y}}{|\psi(X)=y|}
\end{equation}

\begin{equation}
\label{eq:recall}
recall_{y} = \frac{tp_{y}}{|label(X)=y|}
\end{equation}

\subsection{Expert system \label{subsec:ExpertSystem}}
The function $\omega$ gives a lot of information about objects but it needs the instance to return the best pipeline for that instance which is not available a priori. Indeed, this is what we want to identify. We use the information that would provide $\omega$ to build a hierarchical classification based in a clustering of similar objects.

Since some parts work better with some particular pipelines because of their shape, color or texture, we try to take advantage of this and make clusters of objects that are classified similarly well by each pipeline. For example, two parts that have textures may be better recognized by pipelines that use descriptors like SIFT or SURF rather than non textured parts. We call these clusters typologies. This clustering is made using the algorithm K-means \cite{macqueen1967}, that aims to partition the objects into K clusters (where $K<|Y|$)  in which each object belongs to the cluster with the nearest centroids. The input is a matrix with the instances as rows and for each row the $F_1$ value of each pipeline. The inputs for this algorithm are for each instance an array of the $F_1$ value obtained with every pipeline. The election of a good K may highly vary the result since if almost all the clusters are composed by 1 instance the result would be close to just using $\psi^{**}$. After obtaining the $K$ typologies, the $\psi^*_T$'s (\ref{eq:psi_t}) are calculated, i.e., the best pipeline for each typology.

\begin{equation}
\label{eq:psi_t}
\psi^*_T = \displaystyle \underset{\psi \in \Psi}{\mathrm{argmax}}\,\frac{\displaystyle \sum_{y \in T} {F_1}_y^{\psi}(X,Y)}{|T|}
\end{equation}

\begin{figure}
    \centering
    \includegraphics[width=\textwidth]{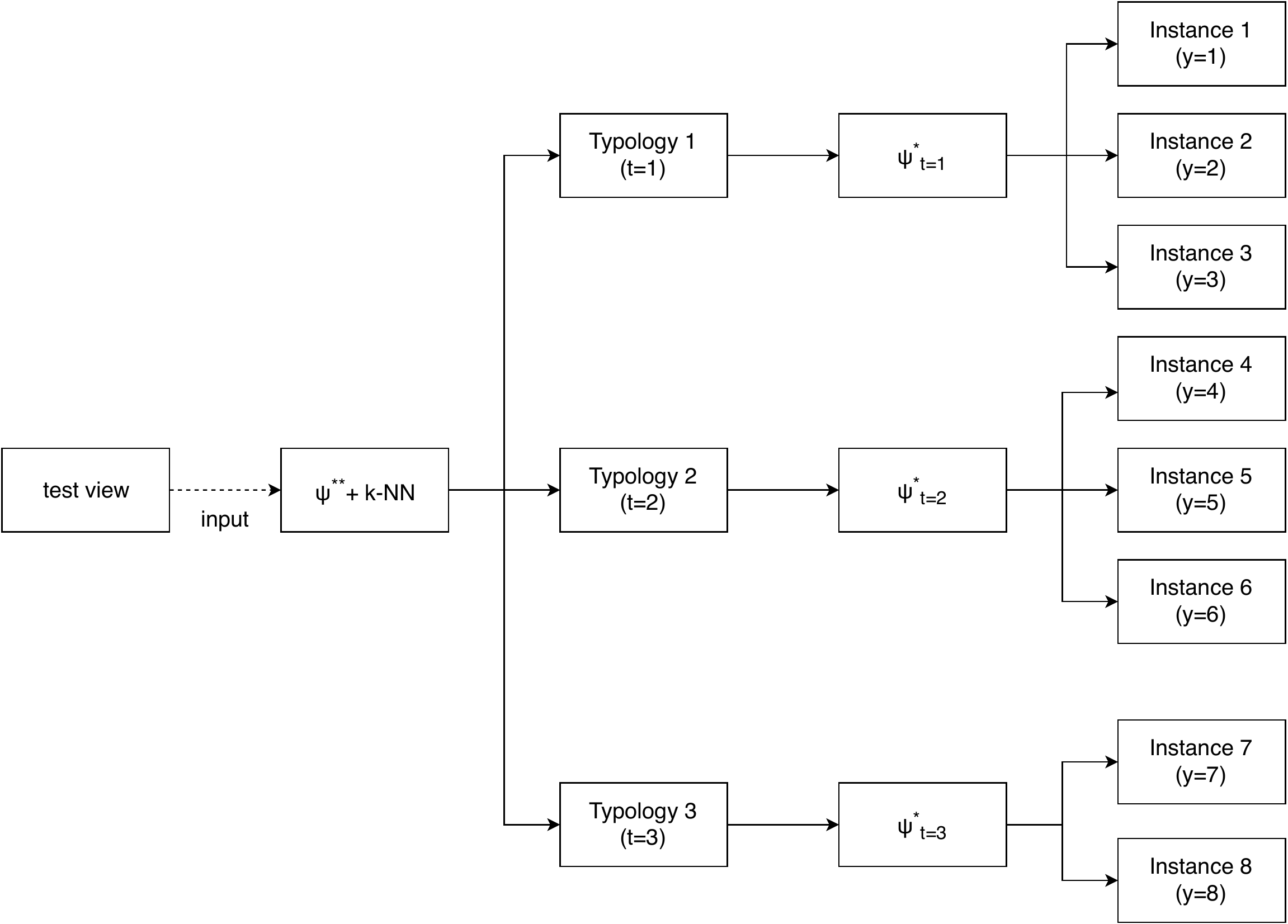}
    \caption{Hierarchical classification}
    \label{fig:hierarchical_clasification}
\end{figure}

The first step of the hierarchical recognition is to recognize the typology with the $\psi^{**}$. Given the typology $t$ as the typology predicted, the $\psi_{t}^{*}$ is used to recognize the instance $y$ of the object. We call the hierarchical recognition $\Upsilon$. The Figure \ref{fig:hierarchical_clasification} shows an scheme of the hierarchical recognition for clarification. 

\section{Experiments and results \label{sec:experiment}}
Our initial hypothesis is that $\Upsilon$ has a better performance than $\psi^{**}$. In order to demonstrate this hypothesis we conducted some experiments. Moreover, we want to know in which way does the number of parts and the number of views per part affect the result. 

The pipelines used (detector, descriptor and matcher) are defined in Subsection \ref{subsec:pipelines}. In Subsection \ref{subsec:ourdataset}, we explain the dataset we have created to evaluate the proposed method under the use case that is the industrial area and the results obtained. In order to compare these results with a well-known dataset in Subsection \ref{subsec:caltech} we present the Caltech dataset \cite{FEIFEI200759} and the results obtained.

\subsection{Pipelines \label{subsec:pipelines}}
The pipelines we have selected are shown in Table \ref{table:pipelines}. Many combination could be done but it is not consistent to match binary descriptors with a L2 distance. The combinations chosen are compatible and may not be the best combination. LBP does not need a detector because it is a global descriptor. 

\begin{table}[]
\centering
\begin{tabular}{|c||c|c|c|}
\hline
Pipeline   & Detector      & Descriptor        & Matcher                      \\ \hline \hline
$\psi_0$        & SIFT          & SIFT              & FLANN                        \\ \hline
$\psi_1$        & SURF          & SURF              & FLANN                        \\ \hline
$\psi_2$         & ORB           & ORB               & \begin{tabular}[c]{@{}l@{}}Brute force\\ Hamming\end{tabular}          \\ \hline
$\psi_3$         & ----          & LBP               & FLANN                        \\ \hline
$\psi_4$       & SURF          & BRIEF             & \begin{tabular}[c]{@{}l@{}}Brute force\\ Hamming\end{tabular}          \\ \hline
$\psi_5$       & BRISK         & BRISK             & \begin{tabular}[c]{@{}l@{}}Brute force\\ Hamming\end{tabular}          \\ \hline
$\psi_6$       & AGAST         & DAISY             & FLANN                        \\ \hline
$\psi_7$       & AGAST         & FREAK             & \begin{tabular}[c]{@{}l@{}}Brute force\\ Hamming\end{tabular}          \\ \hline
\end{tabular}
\caption{Pipelines composition.}
\label{table:pipelines}
\end{table}

\subsection{Our dataset \label{subsec:ourdataset}}
\begin{figure}
    \centering
    \includegraphics[width=\textwidth]{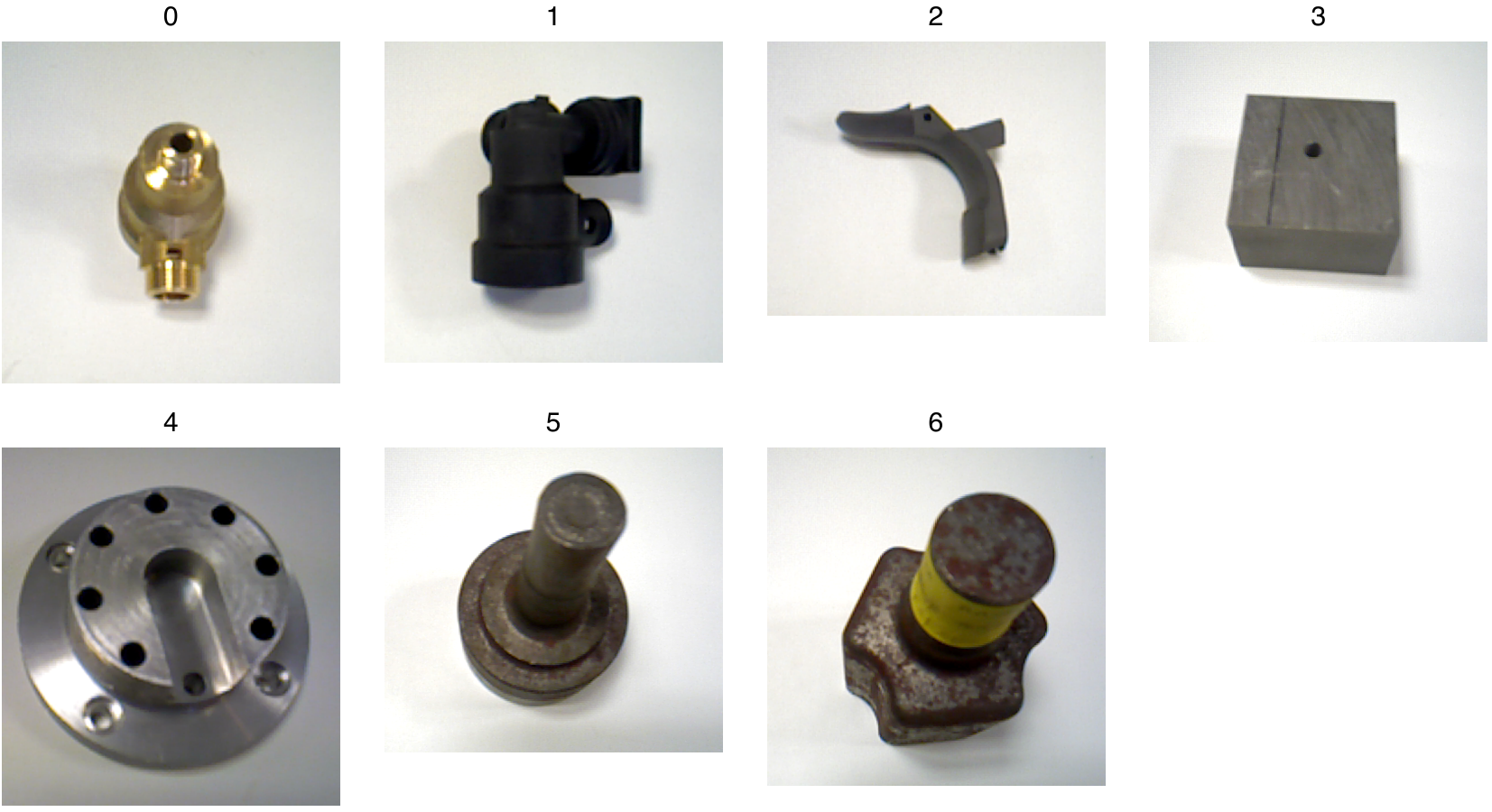}
    \caption{Parts used in our dataset.}
    \label{fig:pieces}
\end{figure}

We select 7 random industrial parts and on a white background we make 50 pictures per part from different angles randomly. That way, we have a dataset with 350 pictures. In Figure \ref{fig:pieces} are shown zoomed in examples of the pictures taken to the parts. 

We use subsets of the dataset to evaluate if changing the number of views per instance and the number of instance vary the performance. This subsets have from 3 to 7 parts and from 10 to 50 views (10 views step). In Table \ref{table:comparative_f1s_ourdataset} are gathered the results for all the subsets using $\psi^{**}$ and $\Upsilon$. The highest score for each subset is in bold. On average the hierarchical recognition performs better. The more parts or views per part, the better that performs the hierarchical recognition comparing with the best pipeline.

\begin{table}[]
\centering
\begin{tabular}{|cc|c|c|c|c|c|c|c|c|c|c|}
\hline
\backslashbox[8mm]& \begin{tabular}[c]{@{}c@{}}$t$\end{tabular} & \multicolumn{2}{c|}{10}&\multicolumn{2}{c|}{20}& \multicolumn{2}{c|}{30}& \multicolumn{2}{c|}{40}&\multicolumn{2}{c|}{50}\\ \cline{3-12}
\begin{tabular}[c]{@{}c@{}}$p$\end{tabular} &\backslashbox[8mm]& $\psi^{**}$& $\Upsilon$& $\psi^{**}$& $\Upsilon$& $\psi^{**}$& $\Upsilon$& $\psi^{**}$& $\Upsilon$& $\psi^{**}$& $\Upsilon$\\ \hline
\multicolumn{2}{|c|}{3}                                                                                                 & \textbf{0.935} & 0.862          & 0.967          & \textbf{0.983} & 0.989          & \textbf{1}     & 0.992          & \textbf{1}     & 0.993         & \textbf{1}     \\ \hline
\multicolumn{2}{|c|}{4}                                                                                                 & \textbf{0.899} & 0.854          & 0.924          & \textbf{0.962} & 0.932          & \textbf{0.966} & \textbf{0.944} & 0.801          & \textbf{0.91} & 0.865          \\ \hline
\multicolumn{2}{|c|}{5}                                                                                                 & \textbf{0.859} & 0.843          & \textbf{0.868} & 0.863          & \textbf{0.883} & 0.818          & 0.876          & \textbf{0.901} & 0.87          & \textbf{0.912} \\ \hline
\multicolumn{2}{|c|}{6}                                                                                                 & 0.865          & \textbf{0.967} & 0.873          & \textbf{0.992} & \textbf{0.891} & 0.87           & 0.88           & 0.88           & 0.856         & \textbf{0.901} \\ \hline
\multicolumn{2}{|c|}{7}                                                                                                 & 0.872          & \textbf{0.9}   & 0.886          & \textbf{0.986} & \textbf{0.894} & 0.891          & 0.88           & \textbf{0.876} & 0.845         & \textbf{0.94}  \\ \hline
\end{tabular}
\caption{$F_1$'s of the $\psi^{**}$'s and $\Upsilon$ for each subset of our dataset. $p$ stands for number of parts and $t$ for number of pictures per part.}
\label{table:comparative_f1s_ourdataset}
\end{table}

\begin{figure*}
    \centering
    \includegraphics[width=\textwidth]{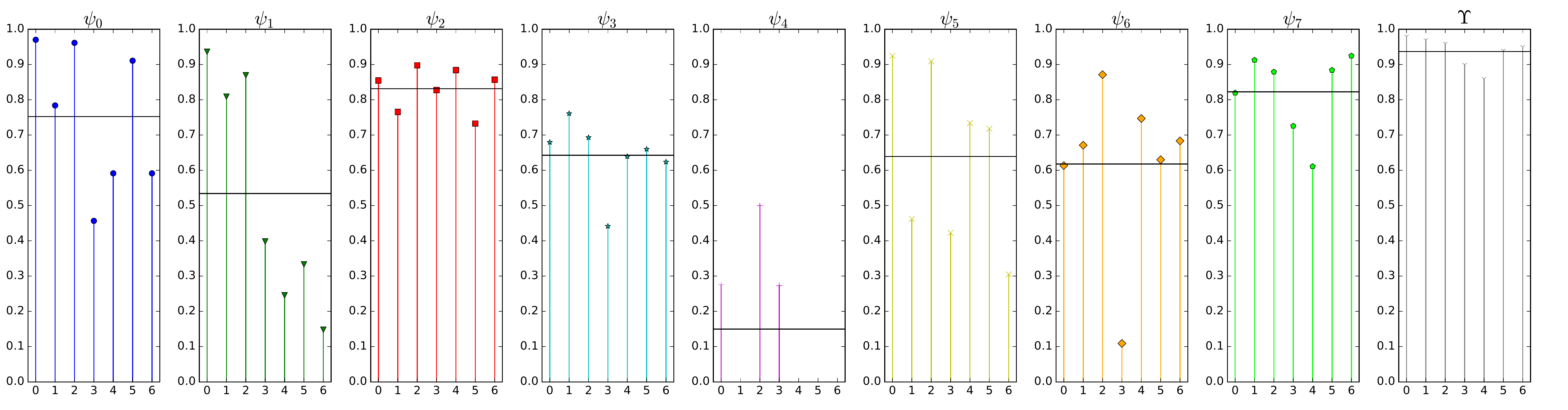}
    \caption{$F_1$ score for each instance and algorithm.}
    \label{fig:f1s_ourdataset}
\end{figure*}

Now we focus on the whole dataset. In Figure \ref{fig:f1s_ourdataset} are shown the $F_1$'s of each instance using each pipeline for this particular case. The horizontal lines mark the $\bar{F_1}$ for that pipeline. The score we obtain with our method (last column) is higher (0.94) than the best pipeline which is $\psi_2$ that corresponds to the pipeline that uses ORB (0.845).

\begin{table}[]
\centering
\begin{tabular}{|c|c|c|c|c|c|c|c|c|}
\hline
$\psi_0$ & $\psi_1$ & $\psi_2$ & $\psi_3$ & $\psi_4$ & $\psi_5$ & $\psi_6$& $\psi_7$ &$\Upsilon$\\ \hline
0.276&0.861&0.976&0.001&0.106 & 0.111&0.296&1.099&1.948 \\ \hline

\end{tabular}
\caption{Time in seconds that needs each pipeline in recognize a piece.}
\label{table:times}
\end{table}

A truthful evaluation of the time performance of the hierarchical classificator is a bit cumbersome since it directly depends on the clustering phase and on which are the best pipelines for each cluster. At least, it needs more time than just using a single pipeline. Given $t(\psi)$ the time need by the pipeline $\psi$, the time needed by $\Upsilon$ is approximately $t(\psi^{**}) + t(\psi^{*}_{T'})$ where $T'$ is the typology guessed by the $\psi^{**}$. In Table \ref{table:times} is shown the time in seconds that each pipeline and the $\Upsilon$ need to recognize a view.

\subsection{Caltech-101 dataset \label{subsec:caltech}}
\begin{figure}
    \centering
    \includegraphics[width=\textwidth]{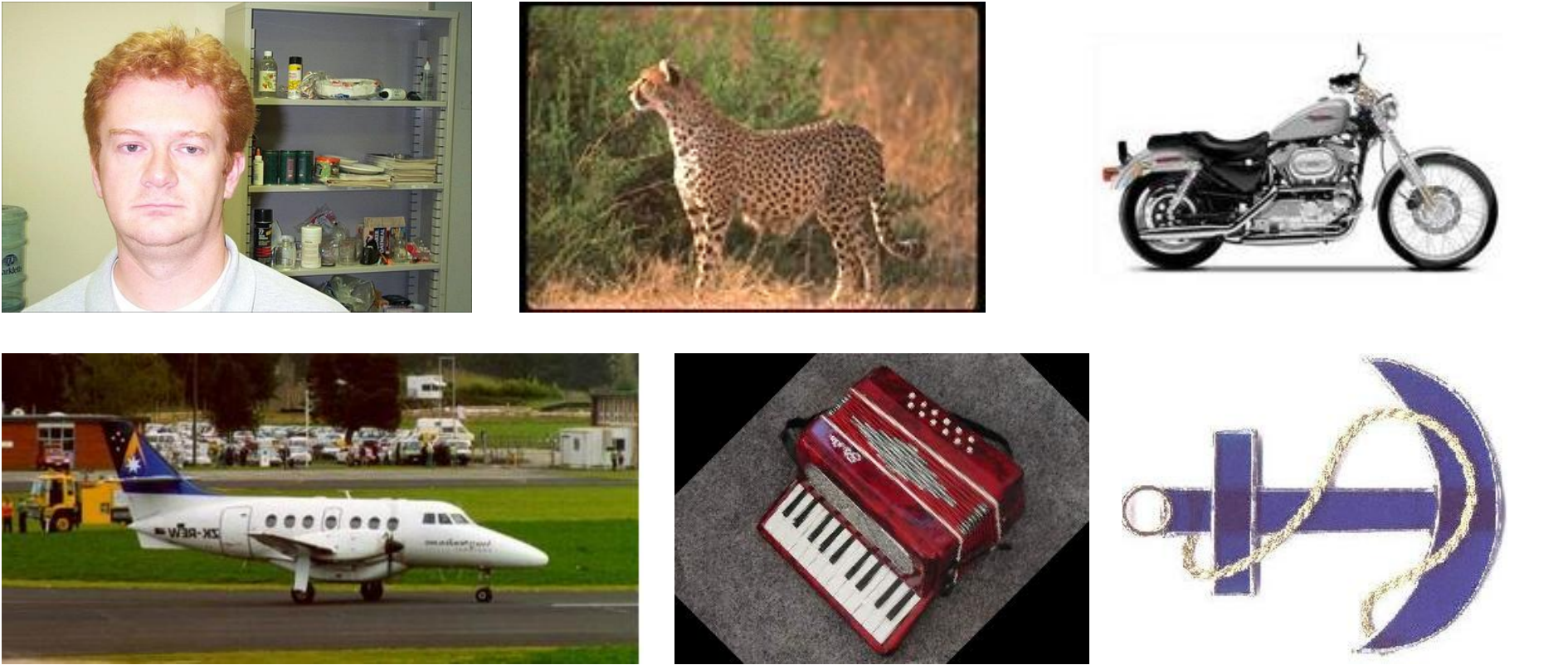}
    \caption{6 random examples of images from the Caltech-101 dataset. The classes are: Face, Leopard, Motorbike, Airplane, Accordion and Anchor.}
    \label{fig:caltech-101}
\end{figure}

Caltech-101 dataset \cite{FEIFEI200759} is a known dataset for object recognition than could be similar to our dataset. This dataset has been tested like our dataset making subsets of the same characteristics. Some randomly picked images from the dataset are shown in Figure \ref{fig:caltech-101}.
{}

The results obtained for the subsets of this datasets are shown in Table \ref{table:comparative_f1s_caltech}. Same conclusions are obtained for this dataset.

\begin{table}[]
\centering
\begin{tabular}{|cc|c|c|c|c|c|c|c|c|c|c|}
\hline
\backslashbox[8mm]& \begin{tabular}[c]{@{}c@{}}$t$\end{tabular} & \multicolumn{2}{c|}{10}&\multicolumn{2}{c|}{20}& \multicolumn{2}{c|}{30}& \multicolumn{2}{c|}{40}&\multicolumn{2}{c|}{50}\\ \cline{3-12}
\begin{tabular}[c]{@{}c@{}}$p$\end{tabular} &\backslashbox[8mm]& $\psi^{**}$& $\Upsilon$& $\psi^{**}$& $\Upsilon$& $\psi^{**}$& $\Upsilon$& $\psi^{**}$& $\Upsilon$& $\psi^{**}$& $\Upsilon$\\ \hline
\multicolumn{2}{|c|}{3}                                                                                                 & 0.967     & 0.967       & 0.983  & 0.983          & 0.989  & 0.989          & 0.992  & 0.992          & 0.993  & 0.993          \\ \hline
\multicolumn{2}{|c|}{4}                                                                                                 & 0.975     & 0.975       & 0.987  & 0.987          & 0.975  & 0.975          & 0.969  & 0.969          & 0.963  & \textbf{0.97}  \\ \hline
\multicolumn{2}{|c|}{5}                                                                                                 & 0.98      & 0.98        & 0.99   & 0.99           & 0.967  & 0.967          & 0.96   & 0.96           & 0.956  & 0.956          \\ \hline
\multicolumn{2}{|c|}{6}                                                                                                 & 0.883     & 0.883       & 0.907  & 0.907          & 0.883  & 0.883          & 0.848  & 0.848          & 0.851  & \textbf{0.936} \\ \hline
\multicolumn{2}{|c|}{7}                                                                                                 & 0.776     & 0.776       & 0.794  & \textbf{0.831} & 0.78   & \textbf{0.84}  & 0.768  & \textbf{0.849} & 0.783  & \textbf{0.843} \\ \hline
\end{tabular}
\caption{$F_1$'s of the $\psi^{**}$'s for each test (Caltech-101). $p$ stands for number of parts and $t$ for number of pictures per part. }
\label{table:comparative_f1s_caltech}
\end{table}

\section{Conclusion \label{sec:conclusion}}
We proposed a hierarchical recognition method based in clustering similar behaviour by the recognition pipelines. It has been demonstrated that on average works better than just recognizing with classical feature-based methods achieving high $F_1$ Scores (in the biggest case, 0.94 for our dataset and 0.843 for the Caltech-101).

As we stated, once we recognize a piece we need its pose to tell the robot where to pick it. This has been let for future work. The use of local features enables the possibility to estimate objects pose using methods such as Hough voting schema, RANSAC or PnP. Additionally, including new feature-based methods may lead to better performance or at least more repeatability and scalability of the hierarchical recognition.

\section*{Acknowledgment}
This paper has been supported by the project SHERLOCK under the European Union’s Horizon 2020 Research \& Innovation programme, grant agreement No. 820689.

\bibliographystyle{plainnat}
\bibliography{root}

\begin{thebibliography}{27}
\providecommand{\natexlab}[1]{#1}
\providecommand{\url}[1]{\texttt{#1}}
\expandafter\ifx\csname urlstyle\endcsname\relax
  \providecommand{\doi}[1]{doi: #1}\else
  \providecommand{\doi}{doi: \begingroup \urlstyle{rm}\Url}\fi

\bibitem[Abdel-Hakim and Farag(2006)]{abdel2006csift}
A.~E. Abdel-Hakim and A.~A. Farag.
\newblock Csift: A sift descriptor with color invariant characteristics.
\newblock In \emph{2006 IEEE Computer Society Conference on Computer Vision and
  Pattern Recognition (CVPR'06)}, volume~2, pages 1978--1983. Ieee, 2006.

\bibitem[Alahi et~al.(2012)Alahi, Ortiz, and Vandergheynst]{alahi_freak:_2012}
A.~Alahi, R.~Ortiz, and P.~Vandergheynst.
\newblock {FREAK}: {Fast} {Retina} {Keypoint}.
\newblock In \emph{2012 {IEEE} {Conference} on {Computer} {Vision} and
  {Pattern} {Recognition}}, pages 510--517. IEEE, June 2012.

\bibitem[Bay et~al.(2006)Bay, Tuytelaars, and Van~Gool]{leonardis_surf:_2006}
H.~Bay, T.~Tuytelaars, and L.~Van~Gool.
\newblock {SURF}: {Speeded} {Up} {Robust} {Features}.
\newblock In \emph{Computer {Vision} – {ECCV} 2006}, pages 404--417. 2006.

\bibitem[Calonder et~al.(2010)Calonder, Lepetit, Strecha, and
  Fua]{calonder2010brief}
M.~Calonder, V.~Lepetit, C.~Strecha, and P.~Fua.
\newblock Brief: Binary robust independent elementary features.
\newblock In \emph{European conference on computer vision}, pages 778--792,
  2010.

\bibitem[Fei-Fei et~al.(2007)Fei-Fei, Fergus, and Perona]{FEIFEI200759}
L.~Fei-Fei, R.~Fergus, and P.~Perona.
\newblock Learning generative visual models from few training examples: An
  incremental bayesian approach tested on 101 object categories.
\newblock \emph{Computer Vision and Image Understanding}, 106\penalty0
  (1):\penalty0 59 -- 70, 2007.

\bibitem[Goutte and Gaussier(2005)]{goutte2005probabilistic}
C.~Goutte and E.~Gaussier.
\newblock A probabilistic interpretation of precision, recall and f-score, with
  implication for evaluation.
\newblock In \emph{European Conference on Information Retrieval}, pages
  345--359, 2005.

\bibitem[Heikkilä et~al.(2009)Heikkilä, Pietikäinen, and
  Schmid]{heikkila_description_2009}
M.~Heikkilä, M.~Pietikäinen, and C.~Schmid.
\newblock Description of interest regions with local binary patterns.
\newblock \emph{Pattern Recognition}, 42\penalty0 (3):\penalty0 425--436, March
  2009.

\bibitem[Indyk and Motwani(1998)]{Indyk:1998:ANN:276698.276876}
P.~Indyk and R.~Motwani.
\newblock Approximate nearest neighbors: Towards removing the curse of
  dimensionality.
\newblock In \emph{Proceedings of the Thirtieth Annual ACM Symposium on Theory
  of Computing}, pages 604--613, 1998.

\bibitem[Ke and Sukthankar(2004)]{yan_ke_pca-sift:_2004}
Y.~Ke and R.~Sukthankar.
\newblock {PCA}-{SIFT}: a more distinctive representation for local image
  descriptors.
\newblock In \emph{Proceedings of the 2004 {IEEE} {Computer} {Society}
  {Conference} on {Computer} {Vision} and {Pattern} {Recognition}, 2004. {CVPR}
  2004.}, volume~2, pages 506--513. IEEE, 2004.

\bibitem[Kohavi(1995)]{kohavi1995study}
R.~Kohavi.
\newblock A study of cross-validation and bootstrap for accuracy estimation and
  model selection.
\newblock In \emph{Ijcai}, volume~14, pages 1137--1145. Montreal, Canada, 1995.

\bibitem[Leutenegger et~al.(2011)Leutenegger, Chli, and
  Siegwart]{leutenegger_brisk:_2011}
S.~Leutenegger, M.~Chli, and R.~Y. Siegwart.
\newblock {BRISK}: {Binary} {Robust} invariant scalable keypoints.
\newblock In \emph{2011 {International} {Conference} on {Computer} {Vision}},
  pages 2548--2555, November 2011.

\bibitem[Liao(2010)]{liao_region_2010}
W.~Liao.
\newblock Region {Description} {Using} {Extended} {Local} {Ternary} {Patterns}.
\newblock In \emph{2010 20th {International} {Conference} on {Pattern}
  {Recognition}}, pages 1003--1006, August 2010.

\bibitem[Lowe(1999)]{lowe_object_1999}
D.~G. Lowe.
\newblock Object recognition from local scale-invariant features.
\newblock In \emph{Proceedings of the {Seventh} {IEEE} {International}
  {Conference} on {Computer} {Vision}}, volume~2, pages 1150--1157 vol.2,
  September 1999.

\bibitem[Lowe(2004)]{lowe_distinctive_2004}
D.~G. Lowe.
\newblock Distinctive {Image} {Features} from {Scale}-{Invariant} {Keypoints}.
\newblock \emph{International Journal of Computer Vision}, 60\penalty0
  (2):\penalty0 91--110, November 2004.

\bibitem[MacQueen(1967)]{macqueen1967}
J.~MacQueen.
\newblock Some methods for classification and analysis of multivariate
  observations.
\newblock In \emph{Proceedings of the Fifth Berkeley Symposium on Mathematical
  Statistics and Probability, Volume 1: Statistics}, pages 281--297. University
  of California Press, 1967.

\bibitem[Mair et~al.(2010)Mair, Hager, Burschka, Suppa, and
  Hirzinger]{mair2010adaptive}
E.~Mair, G.~D. Hager, D.~Burschka, M.~Suppa, and G.~Hirzinger.
\newblock Adaptive and generic corner detection based on the accelerated
  segment test.
\newblock In \emph{European conference on Computer vision}, pages 183--196.
  Springer, 2010.

\bibitem[Mikolajczyk and Schmid(2005)]{mikolajczyk_performance_2005}
K.~Mikolajczyk and C.~Schmid.
\newblock A performance evaluation of local descriptors.
\newblock \emph{IEEE Transactions on Pattern Analysis and Machine
  Intelligence}, 27\penalty0 (10):\penalty0 1615--1630, October 2005.

\bibitem[Muja and Lowe(2009)]{muja2009fast}
M.~Muja and D.~G. Lowe.
\newblock Fast approximate nearest neighbors with automatic algorithm
  configuration.
\newblock \emph{VISAPP}, 2\penalty0 (331-340):\penalty0 2, 2009.

\bibitem[Murala et~al.(2012)Murala, Maheshwari, and
  Balasubramanian]{murala_local_2012}
S.~Murala, R.~P. Maheshwari, and R.~Balasubramanian.
\newblock Local {Tetra} {Patterns}: {A} {New} {Feature} {Descriptor} for
  {Content}-{Based} {Image} {Retrieval}.
\newblock \emph{IEEE Transactions on Image Processing}, 21\penalty0
  (5):\penalty0 2874--2886, May 2012.

\bibitem[Nanni et~al.(2010)Nanni, Brahnam, and Lumini]{nanni2010local}
L.~Nanni, S.~Brahnam, and A.~Lumini.
\newblock A local approach based on a local binary patterns variant texture
  descriptor for classifying pain states.
\newblock \emph{Expert Systems with Applications}, 37\penalty0 (12):\penalty0
  7888--7894, 2010.

\bibitem[Ojala et~al.(1996)Ojala, Pietikäinen, and
  Harwood]{ojala_comparative_1996}
T.~Ojala, M.~Pietikäinen, and D.~Harwood.
\newblock A comparative study of texture measures with classification based on
  featured distributions.
\newblock \emph{Pattern Recognition}, 29\penalty0 (1):\penalty0 51--59, January
  1996.

\bibitem[Pietikäinen et~al.(2011)Pietikäinen, Hadid, Zhao, and
  Ahonen]{pietikainen_local_2011}
M.~Pietikäinen, A.~Hadid, G.~Zhao, and T.~Ahonen.
\newblock Local {Binary} {Patterns} for {Still} {Images}.
\newblock In \emph{Computer {Vision} {Using} {Local} {Binary} {Patterns}},
  Computational {Imaging} and {Vision}, pages 13--47. Springer London, 2011.

\bibitem[Robinson(1981)]{Robinson:1981:KSS:582318.582321}
John~T. Robinson.
\newblock The k-d-b-tree: A search structure for large multidimensional dynamic
  indexes.
\newblock In \emph{Proceedings of the 1981 ACM SIGMOD International Conference
  on Management of Data}, SIGMOD '81, pages 10--18, 1981.

\bibitem[Rosten and Drummond(2005)]{rosten_fusing_2005}
E.~Rosten and T.~Drummond.
\newblock Fusing points and lines for high performance tracking.
\newblock In \emph{Tenth {IEEE} {International} {Conference} on {Computer}
  {Vision} ({ICCV}'05) {Volume} 1}, pages 1508--1515 Vol. 2. IEEE, 2005.

\bibitem[Rublee et~al.(2011)Rublee, Rabaud, Konolige, and
  Bradski]{rublee_orb:_2011}
E.~Rublee, V.~Rabaud, K.~Konolige, and G.~Bradski.
\newblock {ORB}: {An} efficient alternative to {SIFT} or {SURF}.
\newblock In \emph{2011 {International} {Conference} on {Computer} {Vision}},
  pages 2564--2571, November 2011.

\bibitem[Tola et~al.(2010)Tola, Lepetit, and Fua]{tola_daisy:_2010}
E.~Tola, V.~Lepetit, and P.~Fua.
\newblock {DAISY}: {An} {Efficient} {Dense} {Descriptor} {Applied} to
  {Wide}-{Baseline} {Stereo}.
\newblock \emph{IEEE Transactions on Pattern Analysis and Machine
  Intelligence}, 32\penalty0 (5):\penalty0 815--830, May 2010.

\bibitem[Wang and He(1990)]{wang1990texture}
L.~Wang and D.C. He.
\newblock Texture classification using texture spectrum.
\newblock \emph{Pattern Recognition}, 23\penalty0 (8):\penalty0 905--910, 1990.

\end{thebibliography}

\end{document}